\documentclass[runningheads]{llncs}

\usepackage{eccv}

\pdfoutput=1

\usepackage{eccvabbrv}

\usepackage{nicefrac}
\usepackage{graphicx}
\usepackage{booktabs}
\usepackage{multirow}
\usepackage{caption}
\usepackage{subcaption}
\usepackage{booktabs}
\usepackage{bm}

\usepackage[accsupp]{axessibility}  

\usepackage{hyperref}

\usepackage{orcidlink}

\begin{document}

\title{WRIM-Net: Wide-Ranging Information Mining Network for Visible-Infrared Person Re-Identification} 

\titlerunning{WRIM-Net: Wide-Ranging Information Mining Network for Re-ID}

\author{Yonggan Wu\inst{1,2,3}\orcidlink{0000-0002-9836-1749} \and
Ling-Chao Meng\inst{3}\orcidlink{0009-0007-1664-9946} \and
Yuan Zichao\inst{3}\orcidlink{0009-0007-0952-8433}  \and
Sixian Chan\inst{2,4}\orcidlink{0000-0001-8916-1174} \and
Hong-Qiang Wang\inst{2}\orcidlink{0000-0003-1734-7855}\thanks{Corresponding author.}
}

\authorrunning{Y.G Wu et al.}

\institute{
University of Science and Technology of China, Hefei 230026, China\\
\email{braverywu@mail.ustc.edu.cn} \and
Institute of Intelligent Machines/Zhongqi AI Joint Lab., HIPS, CAS, Hefei, China \\
\email{hqwang126@126.com} \and
QiXinMingZhi Technology, Hefei 230088, China \and
Zhejiang University of Technology, Hangzhou 310014, China
}

\maketitle

\begin{abstract}
  For the visible-infrared person re-identification (VI-ReID) task, one of the primary challenges lies in significant cross-modality discrepancy. Existing methods struggle to conduct modality-invariant information mining. They often focus solely on mining singular dimensions like spatial or channel, and overlook the extraction of specific-modality multi-dimension information. To fully mine modality-invariant information across a wide range, we introduce the Wide-Ranging Information Mining Network (WRIM-Net), which mainly comprises a Multi-dimension Interactive Information Mining (MIIM) module and an Auxiliary-Information-based Contrastive Learning (AICL) approach. Empowered by the proposed Global Region Interaction (GRI), MIIM comprehensively mines non-local spatial and channel information through intra-dimension interaction. Moreover, Thanks to the low computational complexity design, separate MIIM can be positioned in shallow layers, enabling the network to better mine specific-modality multi-dimension information. AICL, by introducing the novel Cross-Modality Key-Instance Contrastive (CMKIC) loss, effectively guides the network in extracting modality-invariant information. We conduct extensive experiments not only on the well-known SYSU-MM01 and RegDB datasets but also on the latest large-scale cross-modality LLCM dataset. The results demonstrate WRIM-Net's superiority over state-of-the-art methods.
  \keywords{Person Re-identification \and Wide-Ranging \and Cross-modality}
\end{abstract}

\section{Introduction}
\label{sec:intro}
Person re-identification aims to match images of a person of interest in a query set with those in a gallery set, which may have been captured by different cameras\cite{01_ye2021deep}. Existing methods\cite{01_ye2021deep,12_chen2019abd,24_luo2019strong,25_sun2018beyond,26_wang2018learning,32_bai2020deep} have achieved remarkable performance in visible-light person re-identification. However, visible-light cameras face challenges in capturing clear images under poor lighting conditions. To address this deficiency, infrared cameras are widely employed in modern surveillance systems. Based on this, the VI-ReID task has garnered increasing attention. The VI-ReID task is to match individuals in a gallery set who share the same ID but belong to the opposite modality (infrared or visible).

\begin{figure}[t]
	\centering
		\includegraphics[width=0.80 \linewidth]{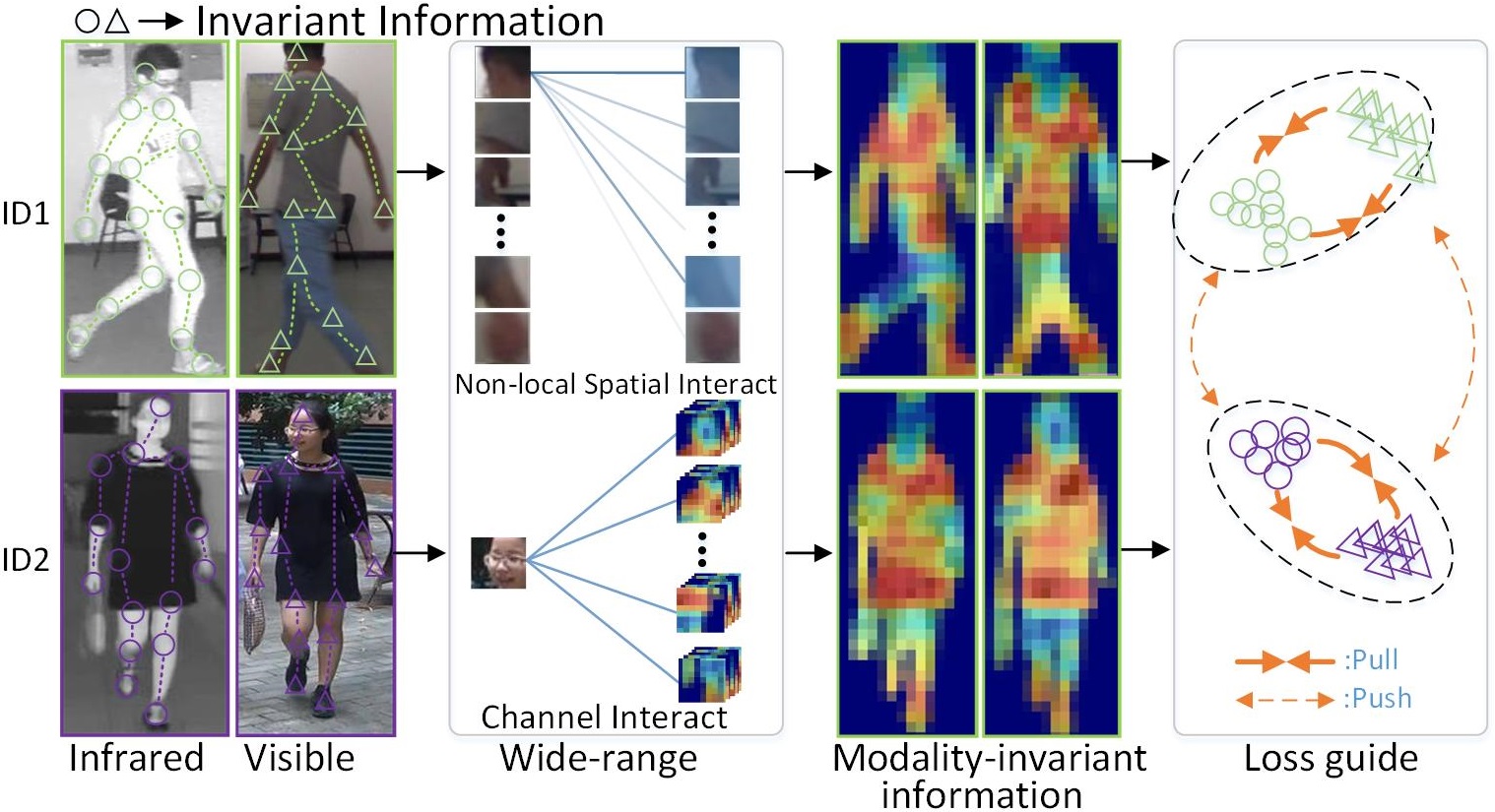}
		\caption{Example of noticeable modality discrepancy between cross-modality images. The motivation behind the proposed WRIM-Net is to mine modality-invariant information across a wide range (e.g. non-local spatial interaction, channel interaction, specific-modality, shared-modality) and to guide the network in better mining invariant information through a novel cross-modality loss.}\label{fig:1}
\end{figure}

VI-ReID is an immensely challenging task, primarily attributed to significant modality discrepancies, as shown in Figure \ref{fig:1}. Existing methods can be broadly categorized into two main types: feature-level methods and image-level methods. Feature-level methods aim to map cross-modality features into a common space\cite{01_ye2021deep,02_wu2021discover,41_ye2020cross,36_chen2021neural}, while image-level methods \cite{49_choi2020hi,34_wang2019rgb,43_wang2020cross,33_wang2019learning} generate diverse-mode images to reduce the modality gap. These methods demonstrate outstanding performance in VI-ReID tasks. However, they still fall short in fully mining modality-invariant information. On one hand, the feature-level methods easily overlook specific-modality multi-dimension information. On the other hand, the image-level methods encounter difficulties in extracting sufficient information due to the scarcity of image pairs.

In Figure \ref{fig:1}, we observe significant differences in images across different modalities. Drawing inspiration from the human visual system, we prioritize the overall shape and the connections between details without focusing on very subtle details per se. For example, in the spatial dimension, our primary focus lies in the non-local spatial relationships, such as contour, posture, and proportions between different body parts (head, arms, knees, etc.). In the channel dimension, due to significant differences in color between modalities, our focus is on information such as texture, which reflects variations in color. All of these constitute modality-invariant information, an aspect where the methods mentioned above are less proficient at mining. Specifically, these methods struggle to simultaneously explore information across multi-dimensions and multi-modalities (specific and shared modality). For example, AGW\cite{01_ye2021deep}, DDAG\cite{06_ye2020dynamic}, and MPANet\cite{02_wu2021discover} employed attention mechanism for extracting information on spatial or channel dimension. However, they applied attention mechanisms in either spatial or channel dimensions, neglecting multi-dimension interaction. Furthermore, they only consider deep layers, disregarding the significance of specific-modality information. While some methods like AGW\cite{01_ye2021deep} proposed a dual-stream architecture to extract specific-modality information, they merely segregate different modalities without specific-modality multi-dimension information mining. 

\begin{figure*}[t]
   \begin{center}
	\includegraphics[width=0.95 \linewidth]{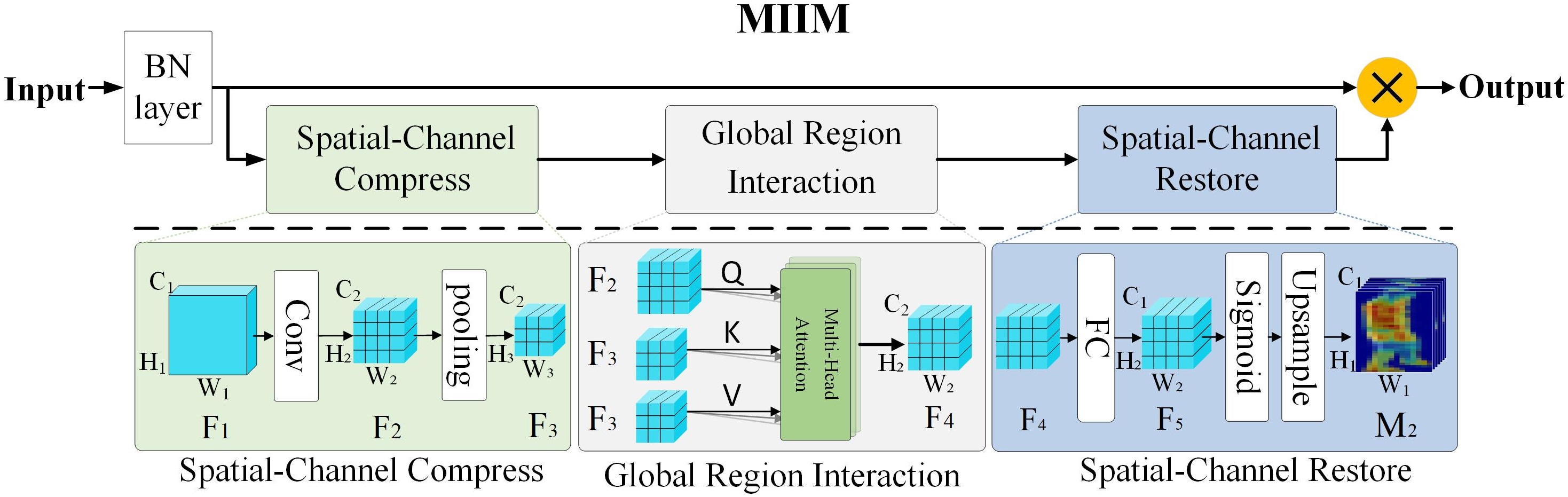}
	\caption{ Diagram of MIIM module. The input features first pass through a standard batch normalization layer and then pass through the Spatial-Channel Compress (SCC) component to compress the size. Subsequently, the feature is passed to the  Global Region Interaction (GRI) component, which employs Multi Head Attention (MHA). Finally, the feature weights are restored to the same size as the input features through the Spatial-Channel Restore (SCR) component. $\otimes$ denotes element-wise multiplication.}
	\label{fig:03} 
    \end{center}
\end{figure*}

Building upon the above issues and aiming to mine modality-invariant information across a wide range, we propose the Multi-dimension Interactive Information Mining (MIIM) module, as illustrated in Figure \ref{fig:03}. Enhanced by the well designed Global Region Interaction (GRI) module, MIIM establishes long-range dependencies through global spatial interactions, effectively extracting non-local spatial-related information such as posture and shape, and simultaneously extracts channel-related information like texture through channel interactions. Furthermore, by being placed in both shallow and deep layers, MIIM captures specific-modality and shared-modality multi-dimension information.

\begin{figure*}[t]
	\begin{center}
		\includegraphics[width=0.80 \linewidth]{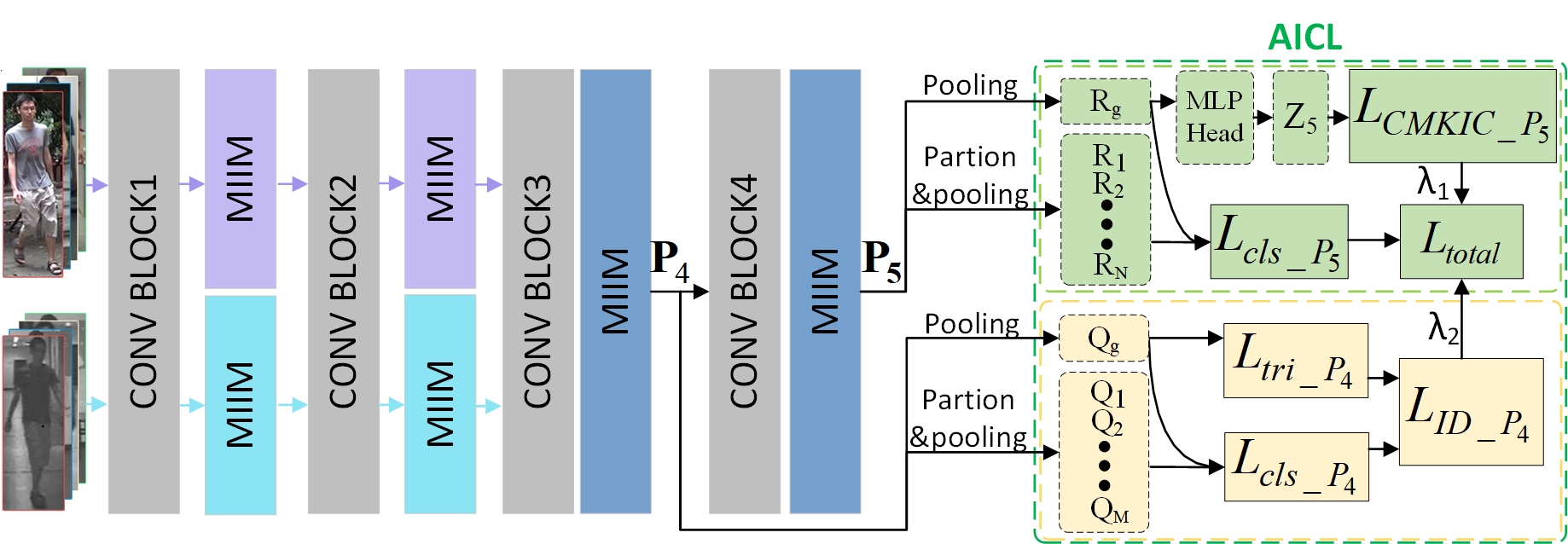}
		\caption{Framework of WRIM-Net. Two separate MIIMs are inserted after each of the first two blocks of the network and a shared MIIM is inserted after each of the last two blocks of the network. Apart from separate MIIM, all other network parameters are shared. AICL uses the traditional ID loss after Block 3 of the network and CMKIC loss after Block 4 of the network. }
		\label{fig:02}
	\end{center}
\end{figure*}

To address the significant modality gap in VI-ReID tasks, guiding the network in better mining modality-invariant information is crucial. \cite{55_wang2020understanding,56_wang2021understanding} suggest that contrastive loss aids the network in mining invariant information by attracting positive pairs and separating negative samples. Building upon the aforementioned considerations, We introduce an Auxiliary-Information-based Contrastive Learning (AICL) approach, depicted in the right portion of Figure \ref{fig:02}, which devises a Cross-Modality Key-Instance Contrastive (CMKIC) loss. The core of AICL lies in employing the CMKIC loss to guide the network in mining modality-invariant information. Additionally, an extra layer of auxiliary information is utilized to further enhance the mining of modality-invariant information.

In summary, our main contributions are summarized below: 

\begin{itemize}
	\item We present the Wide-Ranging Information Mining Network (WRIM-Net). It enables deep information mining across a broad range, encompassing two aspects: (1) Multiple dimensions: spatial and channel-dimension. (2) Multiple modalities: specific-modality and shared-modality.
	\item A plug-and-play Multi-dimension Interactive Information Mining (MIIM) module is designed. It effectively establishes non-local spatial and channel interactions to better mine modality-invariant information.
	\item An Auxiliary-Information-based Contrastive Learning (AICL) approach is devised, employing the novel Cross-Modality Key-Instance Contrastive (CMKIC) loss. AICL leverages auxiliary information and CMKIC loss to more effectively capture modality-invariant information.
	\item Extensive experiments conducted on three benchmark datasets demonstrate the superiority of our model over previous methods. WRIM-Net achieved the best performance of almost all the metrics on all three benchmarks.
\end{itemize}

\section{Related Work}

\subsection{Visible-Infrared Person Re-Identification}

VI-ReID is a challenging task due to the modality differences between infrared and visible images. Existing image-level methods fuse modalities at the data level to alleviate the modality discrepancy~\cite{34_wang2019rgb,42_dai2018cross,43_wang2020cross}, while feature-level methods map visible (VIS) and infrared (IR) features into a common space\cite{01_ye2021deep,02_wu2021discover,41_ye2020cross,36_chen2021neural}. Often, image-level methods suffer from a lack of high-quality VIS-IR image pairs, leading to noise disruption. Feature-level methods are limited due to insufficient modality-invariant information mining. To better capture specific-modality and shared-modality information, the dual-stream network architecture is commonly employed for VI-ReID tasks\cite{01_ye2021deep,28_liu2020parameter,30_ye2018hierarchical,31_ye2019bi}, where the parameters are separated in the shallow layers to obtain specific-modality information and shared in the deep layers to obtain shared-modality information. However, they merely segregate images from different modalities without multi-dimension information interaction. Our proposed WRIM-Net leverages non-local spatial and channel interactions to extract richer long-dependencies information. Additionally, by strategically placing MIIM at different positions, it effectively explores both specific-modality and shared-modality multi-dimension information.

\subsection{Attention Mechanisms}

Attention mechanisms have found widespread application in visual tasks to promote neural network performance by improving visual representation~\cite{09_hu2018squeeze,10_woo2018cbam,11_vaswani2017attention,29_wang2018non}. In the field of person re-identification, attention mechanisms have been incorporated as demonstrated in~\cite{12_chen2019abd,13_xia2019second,01_ye2021deep,29_wang2018non}. To capture long-range spatial relationships, ~\cite{13_xia2019second} introduced second-order non-local attention modules; ~\cite{01_ye2021deep} introduced the non-local~\cite{29_wang2018non} mechanism for attentional feature extraction within the last two blocks of the network. DDAG\cite{06_ye2020dynamic} extracted potent features for cross-modality person re-identification by employing spatial attention modeling on the local features of pedestrians. However, most of the methods primarily employed attention mechanisms in the deep layers of the network, leading to the neglect of mining specific-modality multi-dimension information. The MIIM we propose here effectively mitigates this issue. MIIM achieves non-local spatial interactions through spatial compression and Global Region Interaction (GRI), and channel interactions via channel compression and expansion. The spatial and channel compression of MIIM effectively reduces computational complexity, enabling MIIM to be positioned in shallower layers for specific-modality multi-dimension information mining.

\subsection{Contrastive Learning}

Contrastive learning plays an important role in self-supervised learning~\cite{14_wu2018unsupervised,15_he2020momentum,16_chen2020simple,40_ye2019unsupervised}, allowing the neural network to extract invariant features efficiently. Contrastive learning also has a significant impact on image-text multi-modalities learning\cite{17_radford2021learning,18_li2021align,19_yu2022coca}. \cite{17_radford2021learning} used contrastive learning to perform modality alignment of images and text for cross-modality retrieval. \cite{18_li2021align,19_yu2022coca} employed contrastive learning initially to achieve modality alignment and shared networks for cross-modality feature learning. \cite{57_yang2022augmented} employed joint contrastive learning to acquire color-invariant features in unsupervised VI-ReID tasks. \cite{64_kim2023partmix} introduced a contrastive regularization loss to regularize the model, where positive samples for this contrastive loss involve mixing part descriptors with the same identity. Unlike the previous methods, our CMKIC loss neither employs additional methods to create data-augmented positive samples nor contrasts a single positive for each anchor. Instead, our CMKIC loss introduces a novel approach for selecting positive samples, specifically choosing top-K samples with the same ID as the anchor, different modalities, and the least similarity. This approach increases task difficulty, enabling the network to mine modality-invariant information.

\section{Methodology}

In this section, we provide a detailed description of the proposed method for the VI-ReID task. Our network uses a pre-trained single-stream network (ResNet50~\cite{20_he2016deep}) to extract visible and infrared features. As shown in Figure \ref{fig:02}, the proposed method consists of two main components. (1) {\bf MIIM module}. MIIM enhances its non-local spatial interaction capability by incorporating the spatial compression and GRI module. (2) {\bf AICL approach}. AICL utilizes the CMKIC loss to effectively guide the network in learning modality-invariant information and incorporates auxiliary information to further enhance its extraction.

\subsection{Multi-dimension Interactive Information Mining Module}

MIIM improves information extraction in several ways: (1) Employing global spatial interactions and establishing long-range dependencies to capture non-local spatial-related information, such as shape, pose, and object proportions. (2) Employing channel interactions to capture channel-related information, such as texture. (3) Employing the separate MIIM in the shallow layers to extract specific-modality multi-dimension information (often overlooked by many existing methods) and the shared MIIM in the deep layers to extract shared-modality multi-dimension information. The MIIM module primarily comprises three components: Spatial-Channel Compression (SCC), Global Region Interaction (GRI), and Spatial-Channel Restore (SCR). The input features first pass through a standard Batch Normalization (BN) layer and then proceed to the SCC. In SCC, features undergo two compressions. The first compression is done with a convolutional layer, and the resulting features serve as the Query (Q) for subsequent GRI. The second compression is executed by employing average pooling on the aforementioned features. Then, the resulting features serve as the Key (K) and Value (V) components of GRI. This approach not only achieves non-local spatial interaction but also further reduces computational complexity.

Specifically, the input features $\bm{\mathit{F_1}}\in\mathbb{R}^{H_1\times W_1\times C_1}$ pass through a convolutional layer with a kernel size and stride of $r_s$, resulting in $\bm{\mathit{F_2}}\in\mathbb{R}^{H_2\times W_2\times C_2}$. $\bm{\mathit{F_2}}$ undergoes average pooling with kernel size and stride of $k_s$, yielding  $\bm{\mathit{F_3}}\in\mathbb{R}^{H_3\times W_3\times C_2}$. The resulting $\bm{\mathit{F_2}}$ and $\bm{\mathit{F_3}}$ features are flattened and then fed into the Global Region Interaction (GRI) component, which employs Multi Head Attention (MHA)\cite{11_vaswani2017attention}. In this setup, the flattened $\bm{\mathit{F_2}}$ functions as the Query (Q), while flattened  $\bm{\mathit{F_3}}$ serves as both the Key (K) and Value (V). The resultant features, denoted as $\bm{\mathit{F_4}}$, are obtained from the GRI module's operation, as indicated by the following formula:
\begin{align}
	&{\bm{\mathit{F_4}}} = \operatorname{GRI}(Q={\bm{\mathit{F_2}}}, K={\bm{\mathit{F_3}}}, V={\bm{\mathit{F_3}}}). \label{Formula 1}
\end{align}

The specific implementation of MHA is detailed in \cite{11_vaswani2017attention}, where the number of heads is set to 8.  To better leverage the positional information, we conduct sin-cos positional embedding\cite{11_vaswani2017attention}. In the processing of the GRI component, we utilize higher-resolution $\bm{\mathit{F_2}}$ as the Query (Q), while lower-resolution $\bm{\mathit{F_3}}$ serves as the Key (K) and Value (V), thereby shifting the focus of spatial interaction from interactions among all spatial points to a greater emphasis on long-range spatial regions, enabling global spatial interaction. After passing through the GRI component, the spatial resolution of $\bm{\mathit{F_4}}$ remains the same as that of $\bm{\mathit{F_2}}$. This design further reduces the computational complexity, enabling the placement of this module in the shallower layers of networks.

Next, $C_2$ is restored to the original number of $C_1$ channels using a linear layer $\bm{\mathit{W}}$, similar to Senet \cite{09_hu2018squeeze}, resulting in feature $\bm{\mathit{F_5}}$. Subsequently, a sigmoid activation is applied to the $\bm{\mathit{F_5}}$ to generate weights $\bm{\mathit{M_1}}$:
\begin{align}
	&{\bm{\mathit{M_1}}} = \operatorname{sigmoid}(\bm{\mathit{W}}({\bm{\mathit{F_5}}})). \label{Formula 2}
\end{align}

$\bm{\mathit{M_1}}$ is then reshaped and upsampled to recover the original spatial scale, resulting in the formation of weights $\bm{\mathit{M_2}}\in\mathbb{R}^{H_1\times W_1\times C_1}$. Finally, the input features $\bm{\mathit{F_1}}$ are element-wise multiplied with $\bm{\mathit{M_2}}$ and then passed through the ReLU and BN layers to produce the output feature $\bm{\mathit{F_6}}$:
\begin{align}
	&{\bm{\mathit{F_6}}} = \operatorname{BN}(\operatorname{ReLU}({\bm{\mathit{F_1}}}\otimes {\bm{\mathit{M_2}}})). \label{Formula 3}
\end{align}

In our implementation, we significantly reduce the complexity of the module by compressing the channels and spatial dimensions. In the shallow layers of the network, the features have a larger spatial size, so we primarily focus on spatial dimension compression. Through judicious compression, we not only establish more effective long-range spatial dependencies but also enable MIIM to be conveniently implemented on shallow specific modalities. In the deeper layers of the network, characterized by longer channel lengths, our primary focus is on channel compression. Compressing both ensures that the network maintains a lightweight level of complexity during the following GRI operations. 

To enable MIIM to better assist the network in information mining, we employ separate MIIM for each modality after Block1 and Block2 to perform specific-modality multi-dimension information mining, as shown in Figure \ref{fig:02}. Following Block3 and Block4, a shared MIIM is applied, which allows the network to perform the mining of shared-modality multi-dimension information.

\subsection{Auxiliary-Information-based Contrastive Learning Approach}

In this subsection, we introduce the proposed AICL approach, as illustrated in Figure \ref{fig:02}. The approach involves employing our designed CMKIC loss behind the block4 layer of the network to guide the learning of modality-invariant features. Subsequently, auxiliary information of the block3 layer is leveraged to enhance the mining of modality-invariant information. Our CMKIC loss is different from common contrastive loss. It doesn’t solely use one augmentation of the anchor as the positive sample. Instead, it selects the top-K least similar samples (Key-Instance) with the same ID but different modalities as positive samples. As the anchor and positive samples are from different modalities, this approach facilitates the network’s exploration of crucial modality-invariant information.

Following the approach used in PCB\cite{25_sun2018beyond} and  MGN\cite{26_wang2018learning}, we partition $\bm{\mathit{P_5}}$, the features extracted by Block4, followed by a global average pooling layer, to obtain local features $\bm{\mathit{R_1}}$ to $\bm{\mathit{R_N}}$. Also, another global average pooling layer is directly applied to $\bm{\mathit{P_5}}$ to obtain the global feature $\bm{\mathit{R_g}}$. $\bm{\mathit{R_g}}$ is then passed through a MLP head \cite{16_chen2020simple,39_chen2020improved} to obtain $\bm{\mathit{Z_5}}$, which is formulated as: 
\begin{align}
&{\bm{\mathit{\bm{\mathit{Z_5}}}}}   = \operatorname{\bm{\mathit{{W_2}}}}(\operatorname{ReLU}(\operatorname{BN}(\operatorname{\bm{\mathit{{W_1}}}}(\bm{\mathit{R_g}})))), \label{Formula 4}
\end{align}

where $\bm{\mathit{W_1}}$ and $\bm{\mathit{W_2}}$ represent a linear layer. Then, CMKIC loss is applied to $\bm{\mathit{Z_5}}$.

CMKIC loss serves as the core component enabling AICL to fulfill its mission. The design of our CMKIC loss for WRIM-Net is as follows:
\begin{equation}
	\label{Formula 5}
	\mathcal{L}_{CMKIC\_P_5\_VI} = \sum_{i\in \bm{\mathit{I}_{vis}}}\frac{-1}{\text{top-K}}\sum_{p\in \bm{\mathit{P_{infra}(i)}}}\log_{}\frac{\operatorname{exp}(\nicefrac{\bm{\mathit{z_i}}\cdot \bm{\mathit{z_p}}}{\tau})}{\sum_{a\in \bm{\mathit{A(i)}}}\operatorname{exp}(\nicefrac{\bm{\mathit{z_i}}\cdot \bm{\mathit{z_a}}}{\tau})}, \quad
\end{equation}

The VI in ${\mathcal{L}_{CMKIC\_P_5\_VI}}$  indicates that the anchor represents the Visible  feature, while the other samples involved in the loss correspond to the Infrared light features. $\mathit{I}_{vis}$ denotes the set of visible samples in the current training batch. $\mathit{P_{infra}(i)}$ represents the set of the top-K least similar infrared samples with the same ID as visible sample $i$.
$\bm{z}$ is the feature obtained after applying $L_2$ normalization to the feature $\bm{z_5}$ from Figure \ref{fig:02}. $\bm{z_p}$ refers to the feature $\bm{z}$ corresponding to a sample extracted from the set $\mathit{P_{infra}(i)}$, while $\bm{z_a}$ denotes the feature $\bm{z}$ corresponding to a sample extracted from the set $\bm{A(i)}$. Here, $\bm{A(i)}$ represents the set of all samples in the infrared modality with different ID from $i$, as well as all samples in $P_{infra}(i)$. 
Similarly, for the case where the anchor is the infrared feature, the corresponding CMKIC loss is ${\mathcal{L}_{CMKIC\_P_5\_IV}}$ (See supplementary materials for the formula). And the final $\mathcal{L}_{CMKIC\_P_5}$ is:

\begin{equation}
	\label{Formula 5-2}
	\mathcal{L}_{CMKIC\_P_5} = \frac{1}{2}(\mathcal{L}_{CMKIC\_P_5\_IV}+\mathcal{L}_{CMKIC\_P_5\_VI}).
\end{equation}

In addition, the global and local features are also jointly optimized using a cross-entropy classification loss as:

\begin{equation}
	\mathcal{L}_{cls\_P_5} = \frac{\left(\operatorname{E}(-\log_{}\operatorname{\mathit{p}}(\bm{\mathit{R_g}}))+ \sum_{i=1}^{N}\operatorname{E}(-\log_{}\operatorname{\mathit{p}}(\bm{\mathit{R_i}}))\right)}{N+1}, \label{Formula 6}
\end{equation}

where $p()$ is the probability that the feature is correctly predicted by the classifier and $E$ represents the expectation. 

On the other hand, to fully leverage the auxiliary information from the additional layer, we use the ID loss to guide the features generated by Block3, $\bm{\mathit{P_4}}$. Similar to those for $\bm{\mathit{P_5}}$, we partition $\bm{\mathit{P_4}}$ and obtain global features denoted as $\bm{\mathit{Q_g}}$ and local features denoted as $\bm{\mathit{Q_1}}$ to $\bm{\mathit{Q_M}}$. We apply the triplet loss\cite{69_schroff2015facenet} to the global features $\bm{\mathit{Q_g}}$ and cross-entropy loss for both the global features $\bm{\mathit{Q_g}}$ and the local features. Finally, the ID loss is written as: 
\begin{align}
	&\mathcal{L}_{ID\_P_4}  = \mathcal{L}_{cls\_P_4} + \mathcal{L}_{tri\_P_4}. \label{Formula 7} 
\end{align}

The loss function $\mathcal{L}_{cls\_P_4}$ is similar to $\mathcal{L}_{cls\_P_5}$. The detailed information for $\mathcal{L}_{cls\_P_4}$ and $\mathcal{L}_{tri\_P_4}$ can be found in supplementary materials.

\subsection{Training and Inference}

During training, we consider the three losses, CMKIC, the cross-entropy classification loss, and the ID loss, and have the total loss as a weighted sum:

\begin{equation}
	\mathcal{L}_{total} =\mathcal{L}_{cls\_P_5} + \lambda_1\mathcal{L}_{CMKIC\_P_5} + \lambda_2\mathcal{L}_{ID\_P_4}, \label{Formula 8}
\end{equation}

where the hyper-parameters $\lambda_1$ and $\lambda_2$ are used to balance the contribution of each loss function. During the testing phase, we combine $\bm{\mathit{R_g}}$, $\bm{\mathit{R_1}}$ to $\bm{\mathit{R_N}}$, $\bm{\mathit{Q_g}}$ and $\bm{\mathit{Q_1}}$ to $\bm{\mathit{Q_M}}$ to form the final features for inference.

\section{Experiments}

\subsection{Datasets and Evaluation Setting}

{\bf SYSU-MM01}~\cite{22_wu2017rgb} dataset comprises 491 identities captured by four visible (VIS) cameras and two infrared (IR) cameras, including two modes: All-Search and Indoor-Search. The training set comprises 22,258 visible images and 11,909 infrared images of 395 unique individuals. The test set contains 96 IDs and uses 3,803 infrared images for the query set. 

{\bf RegDB}~\cite{23_nguyen2017person} dataset consists of 412 pedestrians, each of which has 10 visible-light images and 10 infrared images. It randomly divides images of 412 individuals into two equal parts to create the training and test sets. RegDB has two test modes, including the VIS2IR setting and the IR2VIS setting.

{\bf LLCM}~\cite{59_zhang2023diverse} dataset is the latest VI-ReID benchmark, which comprises 30,921 images from 713 unique identities. The testing set includes 13, 909 images from 351 different identities. Both VIS2IR and IR2VIS modes are employed for evaluating the performance of the VI-ReID. Compared to RegDB and SYSU-MM01, LLCM poses a greater challenge, primarily in three aspects: (1) more images from various perspectives. (2) complex low-light conditions. (3) a longer time span. These collectively make the VI-ReID task considerably more challenging.

{\bf Evaluation metrics:} To make fair comparisons, all experiments in this study use two commonly used metrics to evaluate the performance: Rank-1 accuracy and mean average precision (mAP). 

\subsection{Implementation details}

We chose the pre-trained ResNet50 as the backbone. The BNNeck~\cite{24_luo2019strong} was used in the classification head, and the features of the last layer were split according to the ideas of PCB~\cite{25_sun2018beyond} and MGN~\cite{26_wang2018learning}. For SYSU-MM01 and LLCM, the number of split local features is $N = 2$ and $M = 0$, respectively. For RegDB, $N = 6$ and $M = 2$. The image size is resized to $384\times144$. The remaining implementation details can be found in the supplementary materials.

\begin{table*}[t]
	\begin{center}
		\caption{Comparison of Rank-1 (\%) and mAP (\%) performances with the state-of-the-art methods on SYSU-MM01 and RegDB. R1 is the abbreviation of Rank-1. Red denotes the optimal value under the current metric, while blue represents the suboptimal value.}
		\label{table:1}
		\begin{tabular}{ l|@{}|c c|c c|@{}|c c|c c|@{}|c c|c c }
			\hline 
			&            \multicolumn{8}{|c|@{}|}{SYSU-MM01}      & \multicolumn{4}{|c }{RegDB}\\
			\cline{2-13}
		    \multirow{3}{*}{Methods}       &\multicolumn{4}{|c|@{}|}{All-Search}&\multicolumn{4}{|c|@{}|}{Indoor-Search}&\multicolumn{2}{|c|}{\multirow{2}{*}{VIS2IR}}&\multicolumn{2}{|c }{\multirow{2}{*}{IR2VIS}}\\
			\cline{2-9}
			&    \multicolumn{2}{|c|}{Single-shot} & \multicolumn{2}{|c|@{}|}{Multi-shot} & \multicolumn{2}{|c|}{Single-shot} & \multicolumn{2}{|c|@{}| }{Multi-shot}&&&&\\
			\cline{2-13}
			&                                         R1      &     mAP      &      R1      &    mAP      &     R1     &      mAP   &    R1     &   mAP    &    R1   &     mAP &   R1    &     mAP \\
			\hline 
			LbA~\cite{05_park2021learning}         &   55.4    &    57.4      &     57.4     &    59.1     &     -      &       -    &     -     &    -     &  74.2  &  67.6  &  72.4  &  65.5  \\
			HCT~\cite{28_liu2020parameter}         &   61.7    &    57.5      &       -      &      -      &   63.4     &     68.2   &   76.5    &   65.1   &  91.1  &  83.3  &  89.3  &  81.5  \\
			CM-NAS                                 &   62.0    &    60.0      &     68.7     &    53.5     &   67.0     &     73.0   &     -     &    -     &  84.5  &  80.3  &  82.6  &  78.3  \\
                MCLNet~\cite{50_hao2021cross}          &   65.4    &    62.0      &       -      &      -      &   72.6     &     76.6   &   79.6    &   66.6   &  80.3  &  73.1  &  75.9  &  69.5  \\
			SMCL~\cite{37_wei2021syncretic}        &   67.4    &    61.8      &     72.2     &    54.9     &   68.8     &     75.6   &           &          &  83.9  &  79.8  &  83.1  &  78.6  \\
			MPANet~\cite{02_wu2021discover}        &   70.6    &    68.2      &     75.6     &    62.9     &   76.7     &     81.0   &   84.2    &   75.1   &  83.7  &  80.9  &  82.8  &  80.7  \\
			MAUM~\cite{38_liu2022learning}         &   71.7    &    68.8      &       -      &      -      &   77.0     &     81.9   &     -     &     -    &  87.9  &  85.1  &  87.0  &  84.3  \\
			CMT~\cite{04_jiang2022cross}           &   71.9    &    68.6      &     80.2     &    63.1     &   76.9     &     79.9   &   84.9    &   74.1   &  95.2  &  87.3  &  92.0  &  84.5  \\
			MSCLNet~\cite{27_zhang2022modality}    &   77.0    &    71.6      &       -      &      -      &   78.5     &     81.2   &     -     &     -    &  84.2  &  81.0  &  83.9  &  78.3  \\
			SGIEL~\cite{51_feng2023shape}          &   77.1    &    72.3      &       -      &-&\textcolor{blue}{82.1}  &     83.0   &     -     &     -    &  92.2  &  86.6  &  91.1  &  85.2  \\
			PartMix~\cite{64_kim2023partmix}       & \textcolor{red}{\bf77.8}&\textcolor{blue}{74.6}&\textcolor{blue}{80.5}&\textcolor{blue}{69.8} &   81.5    &\textcolor{blue}{84.4}   &   \textcolor{blue}{88.0}   &\textcolor{blue}{80.0}  &  85.7  &  82.3  &  84.9  &  82.5  \\
			DEEN~\cite{59_zhang2023diverse}        &   74.7    &    71.8      &       -      &      -      &   80.3     &     83.3   &     -     &     -    &  91.1  &  85.1  &  89.5  &  83.4  \\
			MUN~\cite{67_yu2023modality}           &   76.2    &    73.8      &       -      &      -      &   79.4     &     82.1   &     -  &-&\textcolor{red}{\bf95.2} &  87.2  &  91.9  &  85.0  \\
			CAL~\cite{68_wu2023learning}           &   74.7    &    71.7      &     77.1     &    64.9     &   79.7     &     83.7   &   87.0   &   78.5  &  94.5  &  \textcolor{blue}{88.7}  &  \textcolor{blue}{93.6}  &  \textcolor{blue}{87.6}  \\
                \hline 
			\bf WRIM-Net                & \textcolor{blue}{77.4}   &  \textcolor{red}{\bf75.4}     &   \textcolor{red}{\bf83.2}    &  \textcolor{red}{\bf71.1}   &  \textcolor{red}{\bf86.2}  &  \textcolor{red}{\bf88.1}   &  \textcolor{red}{\bf92.1} & \textcolor{red}{\bf84.6} &  \textcolor{blue}{94.5}  & \textcolor{red}{\bf90.5} & \textcolor{red}{\bf93.7} & \textcolor{red}{\bf89.7} \\
			\hline 
		\end{tabular}
	\end{center}
\end{table*}

\subsection{Comparison with state-of-the-art methods}

{\bf SYSU-MM01 and RegDB:} Table \ref{table:1} compares our method with SOTA methods. From Table \ref{table:1}, it can be observed that our model achieved the best results by almost all the metrics on both datasets, with only a slight gap in the Rank-1 metric for one mode in RegDB and SYSU-MM01. Specifically, on the SYSU-MM01 dataset, WRIM-Net achieves substantial improvement in the Indoor-Search scenario and Multi-shot mode. 

{\bf LLCM:} Table \ref{table:2} compares WRIM-Net and previous methods on the LLCM dataset. The results show that our model consistently outperforms the state-of-the-art, with all metrics significantly exceeding it.

\subsection{Ablation Study}

The WRIM-Net consists two components: the MIIM module and the AICL approach. In this section, we perform a detailed ablation study to evaluate the role of each component. The experiment was done on the LLCM dataset. The results are shown in Table \ref{table:3}.

\begin{table}[t]
    \begin{minipage}[h]{0.52\textwidth}
        \centering
        \caption{Comparison of Rank-1 (\%) and mAP (\%) with the state-of-the-art on LLCM.}
        \label{table:2}
        \begin{tabular}{ l|c  c|c  c }
            \hline 
            \multirow{3}*{Method}        &\multicolumn{4}{|c }{LLCM} \\
            \cline{2-5}
            &            \multicolumn{2}{|c|@{}|}{IR2VIS} & \multicolumn{2}{|c }{VIS2IR}  \\
            \cline{2-5}
            &              R1      &        mAP     &     R1    &        mAP      \\
            \hline 
            DDAG~\cite{06_ye2020dynamic}       &   41.0    &    49.6    &   48.5    &   53.0     \\
            LbA~\cite{05_park2021learning}     &   44.6    &    53.8    &   50.8    &        55.9     \\
            AGW~\cite{01_ye2021deep}           &   46.4    &    54.8    &   56.0    &   59.1     \\
            CAJ~\cite{61_ye2021channel}        &   48.8    &    56.6    &   56.5    &   59.8     \\
            DART~\cite{62_yang2022learning}    &   52.2    &    59.8    &   60.4    &   63.2     \\
            MMN~\cite{63_zhang2021towards}     &   52.5    &    58.9    &   59.9    &   62.7     \\
            DEEN~\cite{59_zhang2023diverse}    &   54.9    &    62.9    &   62.5    &   65.8     \\
            \hline 
            \bf WRIM-Net                 & \bf58.4   &  \bf64.8   &  \bf67.0  & \bf69.2    \\
            \hline 
        \end{tabular}
    \end{minipage}
    \hfill
    \begin{minipage}[h]{0.42\textwidth}
        \centering
        \caption{Evaluation of different components of the proposed method on LLCM in terms of Rank-1 (\%) and mAP (\%). The mode is IR2VIS.}
        \label{table:3}
        \begin{tabular}{ c|c|c|c|c }
            \hline 
            Baseline  &    MIIM         &      AICL        &      Rank-1      &      mAP     \\
            \hline 
            \checkmark  &   $\times$     &     $\times$      &      50.42       &     57.05    \\
            \hline 
            \checkmark  & \checkmark     &     $\times$      &      55.31       &     61.56    \\
            \hline 
            \checkmark  &   $\times$     &   \checkmark      &      54.83       &     61.27    \\
            \hline 
            \checkmark  & \checkmark     &   \checkmark      &    \bf 58.38       &   \bf 64.75   \\
            \hline 
        \end{tabular}
    \end{minipage}
\end{table}

{\bf Baseline.} The baseline method uses ResNet50 as the backbone network, followed by the BNNeck~\cite{24_luo2019strong}, and finally a fully connected layer as the classifier. The baseline employs both global and local features, guiding $\bm{\mathit{P_5}}$ solely through cross-entropy loss, without auxiliary information of $\bm{\mathit{P_4}}$.

{\bf Effectiveness of MIIM and AICL.}
As shown in Table \ref{table:3}, the results are significantly improved with the addition of MIIM compared to the baseline. This implies that the MIIM module effectively mines modality-invariant information. Table \ref{table:3} also reveals significant improvements in the model's capabilities with the inclusion of AICL. Specifically, introducing MIIM alone results in a $4.89\%$ increase in Rank-1 and a $4.51\%$ increase in mAP. Introducing AICL alone leads to a Rank-1 increase of $4.41\%$ and an mAP increase of $4.22\%$. When both MIIM and AICL are incorporated into the model, the Rank-1 further rises by $7.96\%$, and the mAP increases by $7.7\%$. For further details on the ablation of fusion modes within the MIIM's internal SCR module and auxiliary information in AICL, please refer to the supplementary materials.

{\bf Impact of Different Configurations of MIIM.} The results of different configurations of MIIM are presented in Table \ref{table:4}. 
From Table \ref{table:4}, it can be observed that when employing separate MIIM after Block $1$ and $2$ (Block1/2) alone, the Rank-1 accuracy increases by $2.99\%$, and the mAP increases by $2.76\%$ compared to the configuration without MIIM. Furthermore, utilizing only shared MIIM after Block $3$ and $4$ (Block3/4) yields a $1.33\%$ increase in Rank-1  and a $1.45\%$ improvement in mAP. It is evident that placing MIIM in shallower layers of the network for specific-modality multi-dimension information mining, an aspect overlooked by existing methods, can lead to more significant performance improvements. This is attributed to the innovative and low-complexity design of MIIM.
For more details on the ablation studies of MIIM configurations, please refer to the supplementary materials.

\begin{table}[t]
    \begin{minipage}[t]{0.51\textwidth}
        \centering
        \caption{Impact of different placement of MIIMs on LLCM. The MIIM placed after Block1 and 2 is the separate MIIM, while the MIIM placed after Block3 and 4 is the shared MIIM.}
        \label{table:4}
        \begin{tabular}{ c|c|c|c }
            \hline 
            Block1/2   &   Block3/4      &        Rank-1      &      mAP     \\
            \hline  
            $\times$    &  $\times$      &        54.83       &    61.27    \\              
            \hline  
            \checkmark  &  $\times$      &        57.82       &    64.03    \\  
            \hline
            $\times$    & \checkmark     &        56.16       &    62.72     \\
            \hline 
            \checkmark  & \checkmark     &      \bf58.38      &  \bf64.75    \\
            \hline 
        \end{tabular}        
    \end{minipage}  
  \hfill
  \begin{minipage}[t]{0.41\textwidth}
    \centering
    \caption{Comparison with triplet loss and common contrastive loss on LLCM. "triplet" and "contrastive" respectively represent WRIM-Net replacing CMKIC loss with triplet loss and common contrastive loss.}
    \label{table:7}
    \begin{tabular}{ l|c|c }
        \hline 
                              &       Rank-1     &      mAP     \\
        \hline  
          triplet             &      56.53       &    63.19     \\  
        \hline  
          contrastive         &      56.79       &    63.39     \\     
        \hline  
          WRIM-Net            &   \bf58.38       & \bf64.75     \\    
        \hline 
    \end{tabular}   
   \end{minipage}
\end{table}

\begin{table}[htb]
    \centering
    \caption{Comparison with other attention mechanisms on LLCM. The mode is IR2VIS.}
    \label{table:8}
    \begin{tabular}{ l|c|c|c|c  }
        \hline 
                                            &      flops(B)     &    params(M) &      Rank-1       &      mAP     \\
        \hline  
           CBAM\cite{10_woo2018cbam}         &      6.25         &     26.03    &      50.71        &     57.83     \\  
        \hline  
          Non-Local~\cite{29_wang2018non}     &      7.27         &     27.20    &      50.66        &     57.25     \\     
        \hline  
           MIIM                              &      7.19         &     28.99    &   \bf55.31        &  \bf61.56    \\    
        \hline 
    \end{tabular}
\end{table}

{\bf Comparison with triplet loss and common contrastive loss.} To validate the superiority of the CMKIC loss, we compare it with the triplet loss and common contrastive loss (randomly selecting one positive sample). The experiments were conducted under complete configuration conditions, and the results are presented in Table \ref{table:7}. From the table, we can observe that our CMKIC loss significantly outperforms traditional triplet loss and contrastive loss.

{\bf Comparison with other attention mechanisms.} To demonstrate the superiority of MIIM over other attention methods, we compare MIIM with CBAM\cite{10_woo2018cbam} and Non-Local\cite{29_wang2018non}. CBAM conducts spatial and channel attentions within each block of the backbone network, while Non-Local focuses solely on spatial attention within specific blocks. From Table \ref{table:8}, we can observe that MIIM significantly outperforms CBAM and Non-Local. Here, MIIM refers to the network with the baseline augmented by MIIM. Non-Local and CBAM configurations are applied to the backbone network based on ~\cite{10_woo2018cbam,29_wang2018non}. Our Rank-1 accuracy is $4.60\%$ higher, and mAP is $3.73\%$ higher compared to CBAM. Furthermore, although MIIM demonstrates comparable complexity to Non-Local, it notably outperforms Non-Local.

\subsection{Parameters analysis}
{\bf Analysis of parameters $\lambda_1$ and $\lambda_2$.} Setting $\lambda_1$ and $\lambda_2$ to 0.5 and 0.1 respectively gives the best results. Details can be found in the supplementary materials.

\begin{table}[htb]
    \centering
    \begin{minipage}[b]{0.48 \textwidth}
        \centering
        \caption{Comparison of the results of different spatial compression ratios at each stage on LLCM. The channel compression ratios is set to $2/2/4/4$.}
        \label{Table:11}
        \begin{tabular}{ l|c|c|c }
            \hline 
             $r_{s1}/r_{s2}/r_{s3}/r_{s4}$ &     $8/4/2/2$   &    $4/2/1/1$   &      $2/1/1/1$    \\
            \hline 
               Rank-1      &      57.87       &     \bf58.00     &       57.71\\
            \hline 
               mAP         &      64.17        &    \bf64.32     &       64.04\\
            \hline 
        \end{tabular}
  \end{minipage}
  \hfill
  \begin{minipage}[b]{0.48\textwidth}
        \centering
        \caption{Comparison of the results of different channel compression ratios at each stage on LLCM. The spatial compression ratio is set to $4/2/1/1$.}
        \label{Table:12}
        \begin{tabular}{ l|c|c|c }
            \hline 
       $r_{c1}/r_{c2}/r_{c3}/r_{c4}$ &     $4/4/8/8$   &    $2/2/4/4$   &     $1/1/2/2$    \\
            \hline 
               Rank-1      &      57.51        &       \bf58.00      &      57.45  \\
            \hline 
               mAP         &      63.79        &       \bf64.32      &      63.77  \\
            \hline 
            
        \end{tabular}
    \end{minipage}
\end{table}

{\bf Analysis of spatial and channel compression ratio parameters.} In this section, to enable size compression at various ratios, we configure the input size to be $384\times128$. We compare and analyze the results of different spatial and channel compression ratios at each stage on the LLCM dataset, as shown in Tables \ref{Table:11} and \ref{Table:12}. We denote $r_{s1}/r_{s2}/r_{s3}/r_{s4}$ as the spatial compression ratio and $r_{c1}/r_{c2}/r_{c3}/r_{c4}$ as the channel compression ratio after Block1, Block2, Block3, and Block4 in the MIIM, respectively. From Table \ref{Table:11} and Table \ref{Table:12},  it is evident that the most optimal performance is attained when the spatial compression ratio is configured as $4/2/1/1$ and the channel compression ratio is set to $2/2/4/4$. In both spatial and channel dimensions, performance initially improves before declining as the compression ratio increases. In our view, excessively high compression ratios lead to the loss of critical information, while excessively low ratios impede effective information interaction. In terms of spatial dimensions, appropriate spatial compression enables the network to focus on global relationships, such as the overall shape of the human body.

\begin{table}[h]
  \begin{minipage}[b]{0.39\textwidth}
	\centering
        \caption{Comparison of results with different $k_s$ values in the SCC on LLCM. The mode is IR2VIS.}
        \label{table:9}
        \begin{tabular}{ l|c|c|c|c }
            \hline 
               $k_s$      &      1      &     2     &        3      &     4    \\
            \hline  
              Rank-1     &    57.77     &    58.01  &    \bf58.38   &    57.98 \\  
            \hline  
              mAP        &    64.11     &    64.49  &   \bf64.75    &    64.23  \\     
            \hline  
        \end{tabular}
    \end{minipage}
  \hfill
  \begin{minipage}[b]{0.54\textwidth}
	\centering
        \caption{Comparison of results with different top-K values in CMKIC loss on LLCM, where $'W'$ indicates top-K utilization and $'W/O'$ denotes random selection. The mode is IR2VIS.}
        \label{table:10}
        \begin{tabular}{ l|c|c|c|c|c|c }
            \hline 
              top-K     &    1 $W$    &   2 $W$   &  4 $W$    &  4 $W/O$  &  6 $W$   &  8 $W$ \\
            \hline  
             Rank-1    &    57.33     &   58.04   & \bf58.38  &  56.46    &  57.84   &  57.74  \\  
            \hline  
             mAP       &    63.84    &    64.35   & \bf64.75 &    63.14   &  64.36   &  64.00 \\   
             \hline 
        \end{tabular}
    \end{minipage}
\end{table}

{\bf Analysis of the $k_s$ in SCC.}
To determine the optimal $k_s$ parameter, we conducted parameter experiments. The results, shown in Table \ref{table:9}, indicate that setting $k_s$ to 3 yields the best performance. Specifically, setting $k_s$ to 3 leads to a $0.61\%$ increase in Rank-1 and a $0.64\%$ increase in mAP compared to setting $k_s$ to 1 (equivalent to no compression). This compressed design enables GRI to further facilitate global information interaction, enhancing network performance while maintaining low complexity.

{\bf Analysis of top-K in CMKIC Loss.} 
To validate the effectiveness of top-K and determine the optimal value for CMKIC loss, experiments were conducted as shown in Table \ref{table:10}. The results indicate that the best performance is achieved when top-K is set to 4. In CMKIC loss, if we randomly select top-K positive samples, there is a significant decrease of $1.92\%$ in Rank-1 (as shown in the fifth column). This indicates that our top-K Key-Instance design is effective in assisting the network in mining modality-invariant information.

\subsection{Visualization Analysis}
\begin{figure}[t]
    \centering
    \begin{minipage}[t]{0.48 \textwidth}
        \centering
	  \includegraphics[width=0.95 \linewidth]{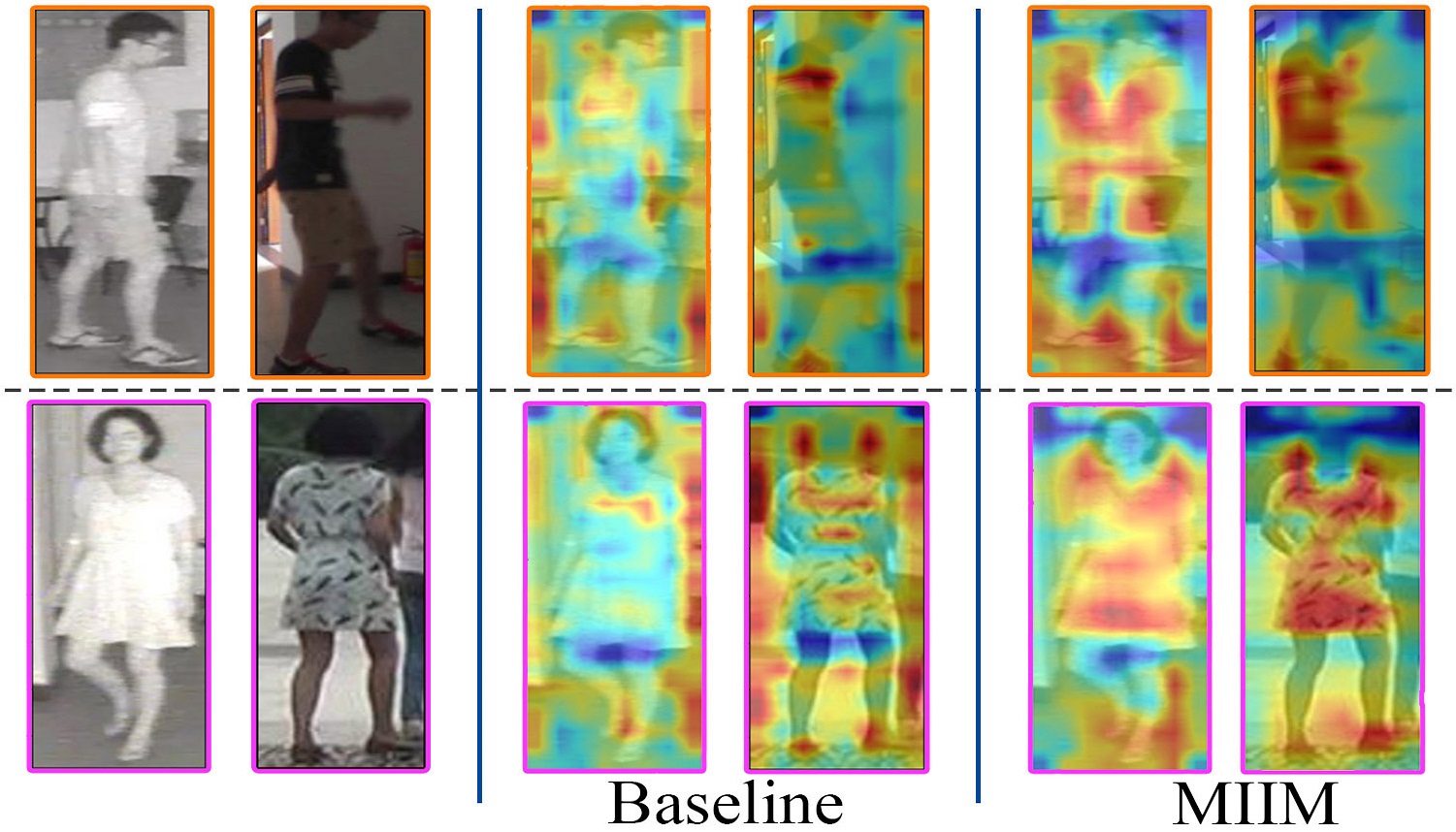}

	\caption{Grad-Cam feature visualization analysis of MIIM. The second column shows the Visualization Heat Map from the baseline network, while the third column displays the Visualization Heat Map with the MIIM module integrated.}
	\label{fig:04}
    \end{minipage}
    \hfill
    \begin{minipage}[t]{0.45 \textwidth}
        \centering
        \includegraphics[width=0.95 \linewidth]{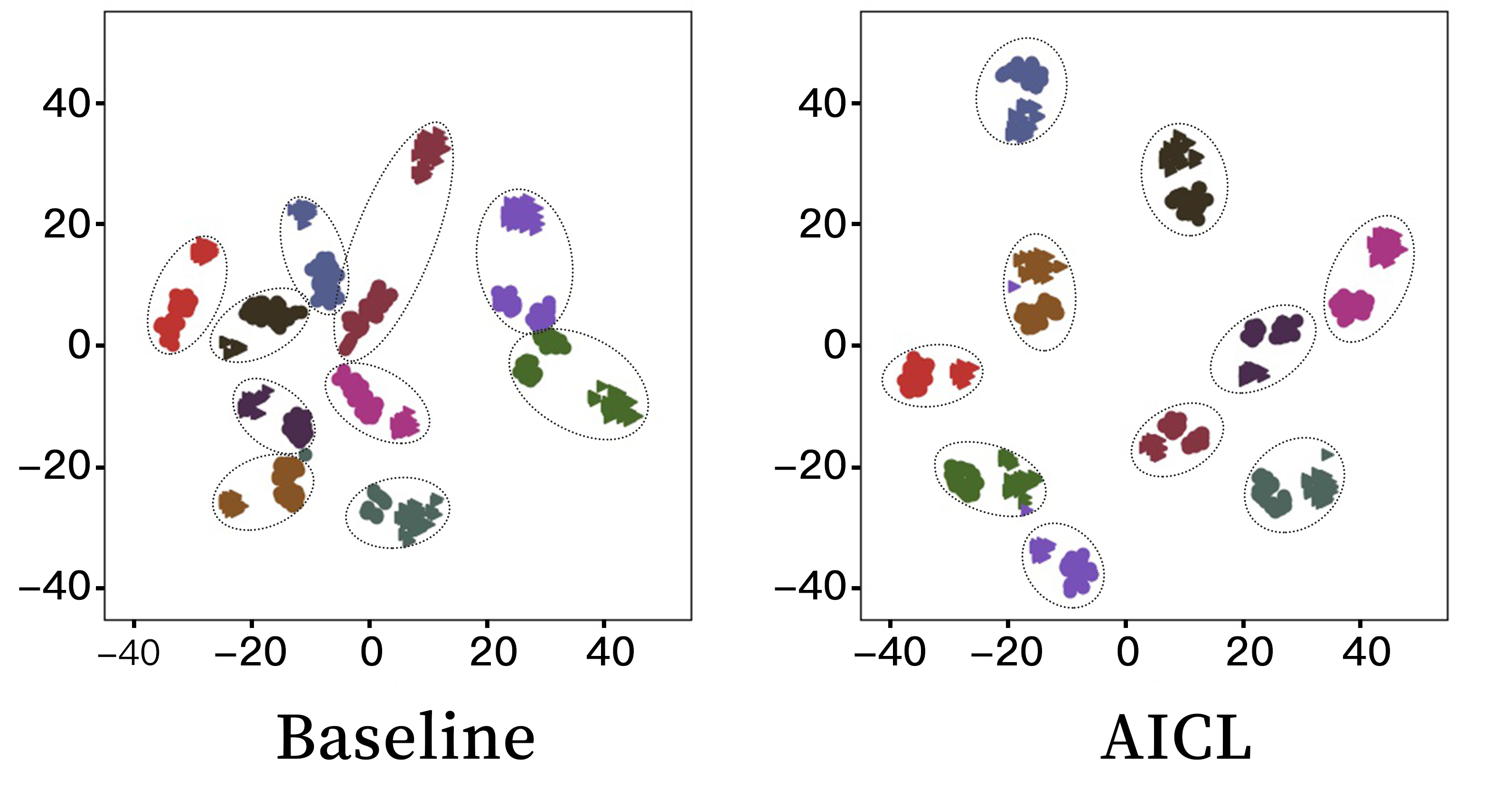}

        \caption{T-SNE feature visualization. Each color represents an ID and the circles and triangles represent different modalities. As can be seen, AICL better alleviates the modality discrepancy and improves the discriminability.}
        \label{fig:05}
    \end{minipage}
\end{figure}

To observe the MIIM module, we utilize Grad-Cam~\cite{46_selvaraju2017grad} to visually analyze the features extracted by both the baseline network and the baseline network added with MIIM. Figure \ref{fig:04} illustrates the visualizations obtained from an infrared query image and its corresponding visible image from the gallery set. From Figure \ref{fig:04}, it can be observed that after adding MIIM, the network effectively focuses on pedestrians themselves, whether in the visible modality or the infrared modality, while paying more attention to global information. We attribute this to MIIM's improved non-local spatial and channel interactions, along with its specific-modality multidimensional information mining.

To observe the AICL, we use T-SNE~\cite{48_van2008visualizing} to visualize and compare the features extracted by the baseline and AICL. As shown in Figure \ref{fig:05}. After using AICL, the feature distance of different modalities of the same ID is pulled closer while the feature distance between different IDs is pushed farther. This allows the network to more effectively mine modality-invariant information.

\section{Conclusion}
We have introduced the WRIM-Net for VI-ReID, which emphasizes modality-invariant information mining across a wide range. To achieve this goal, we developed the MIIM module for extracting specific-modality and shared-modality features through multi-dimension interactions. And as a plug-and-play module, it is placed at different positions of WRIM-Net to expand the range of mining information. We also designed the AICL approach, which guides the network to better explore modality-invariant information by utilizing Cross-Modality Key-Instance Contrastive loss. Extensive experiments on three standard datasets demonstrated the superiority of WRIM-Net with the best performance of almost all the metrics on all three benchmarks.

\par\vfill\par

\clearpage  

\section*{Acknowledgements}
This work was supported in part by the National Natural Science Foundation of China (Grant numbers 61973295, 81872276), Key Research and Development Project of Anhui Province (Grant number 201904a07020092), and Zhejiang Provincial Natural Science Foundation of China (Grant No. LY23F020023)

%
%

\bibliographystyle{splncs04}
\end{document}